\documentclass[10pt,twocolumn,letterpaper]{article}

\usepackage{iccv}              % To produce the CAMERA-READY version
\usepackage{tabularx}
\usepackage{graphicx}
\usepackage{subcaption}
\usepackage{tikz}
\usetikzlibrary{shapes, positioning, backgrounds}
\definecolor{iccvblue}{rgb}{0.21,0.49,0.74}

\usepackage[pagebackref,breaklinks,colorlinks,allcolors=iccvblue]{hyperref}
\def\paperID{*****} % *** Enter the Paper ID here
\def\confName{ICCV}
\def\confYear{2025}

\title{Drone Detection with Event Cameras}

\author{Gabriele Magrini\\
University of Florence\\
Florence\\
{\tt\small gabriele.magrini@unifi.it}
\and
Lorenzo Berlincioni\\
University of Florence\\
Florence\\
{\tt\small lorenzo.berlincioni@unifi.it}
\and
Luca Cultrera\\
University of Florence\\
Florence\\
{\tt\small luca.cultrera@unifi.com}
\and
Federico Becattini\\
University of Siena\\
Siena\\
{\tt\small federico.becattini@unisi.it}
\and
Pietro Pala\\
University of Florence\\
Florence\\
{\tt\small pietro.pala@unifi.it}
}

\begin{document}
\maketitle
\begin{abstract}
The diffusion of drones presents significant security and safety challenges. Traditional surveillance systems, particularly conventional frame-based cameras, struggle to reliably detect these targets due to their small size, high agility, and the resulting motion blur and poor performance in challenging lighting conditions. This paper surveys the emerging field of event-based vision as a robust solution to these problems. Event cameras virtually eliminate motion blur and enable consistent detection in extreme lighting. %, while their sparse data stream naturally filters out static backgrounds to focus on moving objects.
Their sparse, asynchronous output suppresses static backgrounds, enabling low-latency focus on motion cues.
We review the state-of-the-art in event-based drone detection, from data representation methods to advanced processing pipelines using spiking neural networks. The discussion extends beyond simple detection to cover more sophisticated tasks such as real-time tracking, trajectory forecasting, and unique identification through propeller signature analysis. By examining current methodologies, available datasets, and the distinct advantages of the technology, this work demonstrates that event-based vision provides a powerful foundation for the next generation of reliable, low-latency, and efficient counter-UAV systems.
\end{abstract}  

\section{Introduction}
\label{sec:intro}

\begin{figure}[t]
    \centering
    \includegraphics[width=\linewidth]{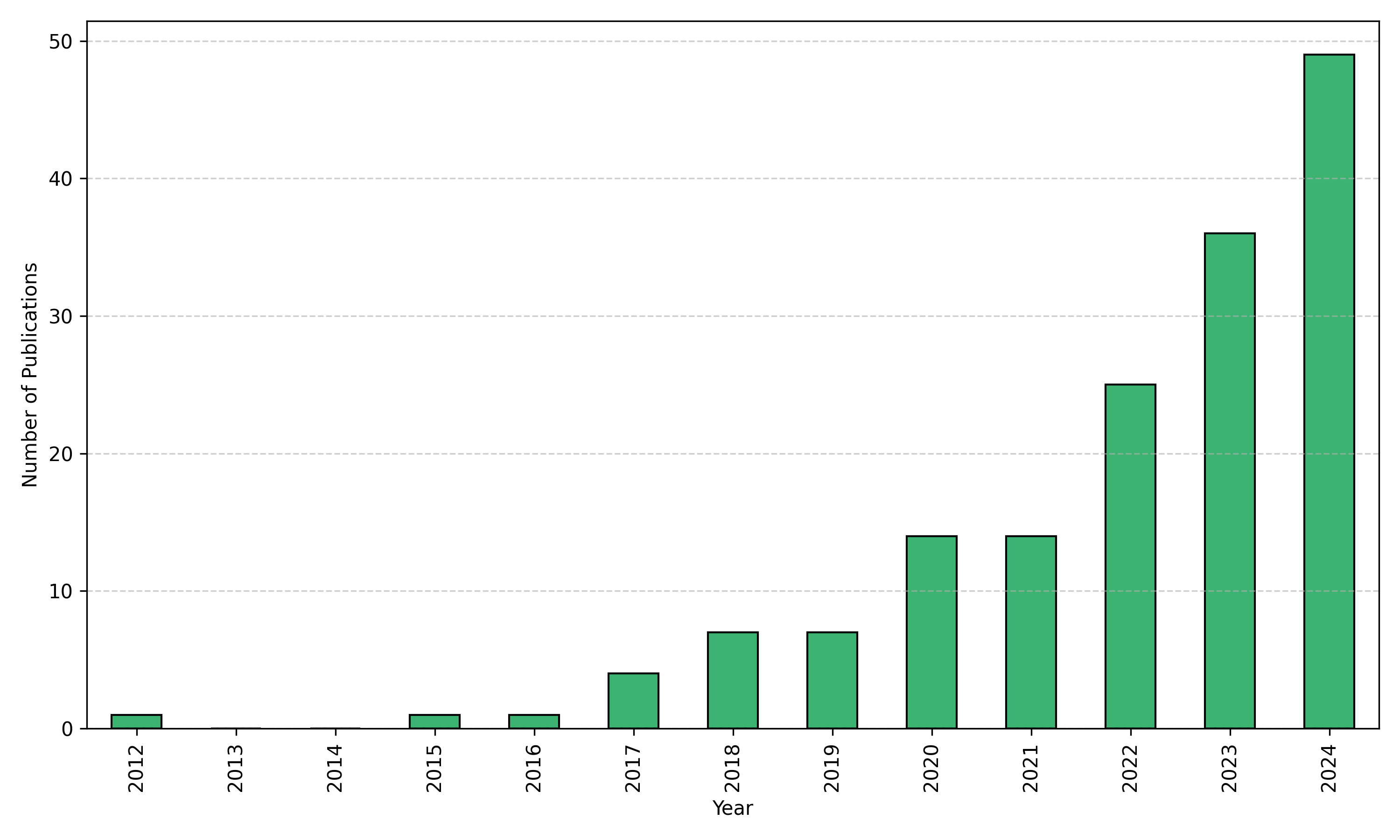}
    \caption{\textbf{Event camera based drone detection works trend.} We show the number of papers trough the years 2012-2024 as results of the query on Google Scholar \texttt{(event\_camera OR neuromorphic\_vision) AND (drone OR UAV) AND (detection)}. }
    \label{fig:num_papers}
\end{figure}

The proliferation of small unmanned aerial vehicles (UAVs), commonly known as drones, has marked a significant technological turning point, unlocking a vast range of applications from commercial logistics and precision agriculture to infrastructure inspection and cinematography. However, this increasing accessibility has concurrently given rise to pressing challenges in security and public safety~\cite{yousaf2022drone}. The potential for malicious use, including unauthorized surveillance, the disruption of controlled airspace, smuggling of contraband, and threats to critical infrastructure, has established the development of robust and reliable counter-UAV systems as a paramount research priority. The intrinsic characteristics of these aircraft, namely their small size, high agility, low acoustic and thermal signatures, and often non-metallic composition, present a serious detection challenge for traditional surveillance technologies.
% \cite{farah2025ev, cao2024eventboost, chuecaonboard, da2025new, magrini2024neuromorphic, stewart2022virtual, mitrokhin2018event, miao2025dual, shu2021small, magrini2025fred, liu2022edflow, iaboni2022event, magrini2025ev, svanstrom2022drone, skogsberg2024event, eldeborg2024drone, han2024event, zhang2024spiking, chen2025event, stewart2021drone, stewart2023using}

Among the various sensing modalities employed for drone detection, vision-based systems are particularly compelling due to their passive nature and ability to provide rich information for target classification and identification~\cite{jiang2021anti,pawelczyk2020real}. Nevertheless, conventional frame-based cameras, which acquire images at a fixed rate, suffer from fundamental limitations when tasked with detecting drone targets. The high angular velocity of a drone, especially in close-range scenarios, frequently leads to significant motion blur, which degrades feature representation and complicates detection and tracking. Furthermore, the limited dynamic range of standard CMOS sensors makes them highly susceptible to challenging lighting conditions. A drone silhouetted against a bright sky can be lost in sensor saturation, while one operating in low-light or shadowed regions may fail to generate sufficient contrast for detection~\cite{muller2017robust, magrini2025fred}. These limitations necessitate a paradigm shift in visual sensing to overcome the specific challenges posed by drone detection.

In response to these shortcomings, a new class of bio-inspired sensors, known as event cameras, has emerged as a technology for high-speed and high-dynamic-range applications \cite{gallego2020event}. Unlike traditional cameras that capture a sequence of intensity frames, event cameras operate asynchronously. Each pixel independently monitors for changes in log-scale brightness. When a change exceeds a programmable threshold, the pixel generates an \textit{event} containing its spatial coordinates, a microsecond-resolution timestamp, and the polarity of the brightness change. This event-driven data acquisition results in a sparse and continuous stream of information that encodes the dynamic aspects of a scene. The key advantages of this paradigm include a temporal resolution on the order of microseconds, which virtually eliminates motion blur; a very high dynamic range of over 120 dB, enabling robust performance in extreme lighting conditions; and low power consumption, as static parts of the scene generate no data.

The unique properties of event cameras are well-suited to the drone detection problem \cite{magrini2024neuromorphic}. The high temporal resolution ensures that the fine details and motion trajectory of a fast-moving drone are captured without blur, providing a rich signal for detection and tracking algorithms \cite{liu2022edflow, mitrokhin2018event}. The high dynamic range allows for consistent detection in scenarios with severe backlighting or poor illumination, where conventional cameras would fail. Furthermore, the inherent data sparsity, which naturally filters out static backgrounds and highlights moving objects, serves as a powerful attentional mechanism. This significantly reduces data redundancy and allows computational resources to be focused on the target of interest, making event cameras an ideal sensor for real-time, low-latency, and power-efficient counter-UAV systems \cite{cao2024eventboost, iaboni2022event}.
In this work, we study these distinct advantages, analyzing the rapid diffusion of event cameras for drone detection (see Fig. \ref{fig:num_papers}).
%demonstrating the significant potential of event-based vision to create the next generation of robust and reliable counter-UAV solutions.
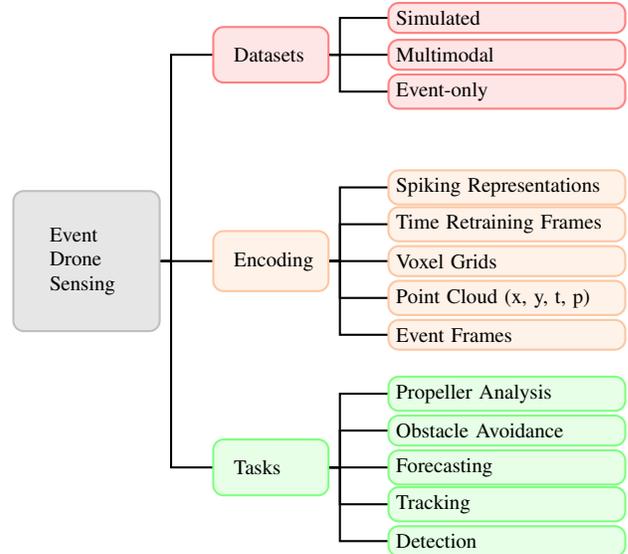
\begin{figure}[t]
\centering
\resizebox{\linewidth}{!}{
\begin{tikzpicture}[
  grow=right,                              % set once
  every node/.style={font=\footnotesize},
  edge from parent/.style={draw, thick},
  level 1/.style={sibling distance=28mm, level distance=25mm},
  level 2/.style={sibling distance=5mm, level distance=32mm},
  level 3/.style={sibling distance=4mm, level distance=38mm,
   edge from parent/.style={draw, thick, ->}},
  edge from parent path={(\tikzparentnode.east) -- ++(4pt,0) |- (\tikzchildnode.west)},
  category/.style={draw=gray!50, fill=gray!20, rectangle, rounded corners, thick, inner sep=14pt,text width=1cm},
  subcat/.style={draw=orange!50, fill=orange!10, rectangle, rounded corners, thick, inner sep=3pt,text width=3.0cm},
  paperbox/.style={draw=gray!60, fill=gray!10, rounded corners, thin, inner sep=2pt, text width=3cm},
  categorypink/.style={draw=Peach!50, fill=Peach!10, rectangle, rounded corners, thick, inner sep=8pt,text width=1cm},
  subcatpink/.style={draw=Peach!50, fill=Peach!10, rectangle, rounded corners, thick, inner sep=3pt,text width=3.0cm},
  categoryred/.style={draw=red!50, fill=red!10, rectangle, rounded corners, thick, inner sep=8pt,text width=1cm},
  subcatred/.style={draw=red!50, fill=red!10, rectangle, rounded corners, thick, inner sep=3pt,text width=3.0cm},
  subcatgreen/.style={draw=green!50, fill=green!10, rectangle, rounded corners, thick, inner sep=3pt,text width=3.0cm},
  categorygreen/.style={draw=green!50, fill=green!10, rectangle, rounded corners, thick, inner sep=8pt,text width=1cm},
  categoryblue/.style={draw=blue!50, fill=blue!10, rectangle, rounded corners, thick, inner sep=8pt,text width=1cm},
  subcatblue/.style={draw=blue!50, fill=blue!10, rectangle, rounded corners, thick, inner sep=3pt,text width=3.0cm},
]

\node[category]{Event\\Drone\\Sensing}
  child { node[categorygreen]{Tasks}
    child { node[subcatgreen]{Detection}
      % child { node[paperbox]{\cite{magrini2025fred}, \cite{mandula2024towards}, \cite{stewart2022virtual}} }
    }
    child { node[subcatgreen]{Tracking}
      % child { node[paperbox]{\cite{iaboni2022event}, \cite{skogsberg2024event}, \cite{wu2017vision}, \cite{mitrokhin2018event}, \cite{magrini2025fred}} }
    }
    child { node[subcatgreen]{Forecasting}
      % child { node[paperbox]{\cite{magrini2025fred}, \cite{falanga2020dynamic}} }
    }
    child { node[subcatgreen]{Obstacle Avoidance}
      % child { node[paperbox]{\cite{bhattacharya2025monocular}, \cite{zsedrovits2011collision}} }
    }
    child { node[subcatgreen]{Propeller Analysis}
      % child { node[paperbox]{\cite{stewart2021drone}, \cite{zhang2024spiking}, \cite{spetlik2025efficient}} }
    }
  }
  child { node[categorypink]{Encoding}
    child { node[subcatpink]{Event Frames}
      % child { node[paperbox]{\cite{eldeborg2024drone}, \cite{magrini2025fred}} }
    }
    child { node[subcatpink]{Point Cloud (x, y, t, p)}
      % child { node[paperbox]{\cite{chen2025event}, \cite{magrini2025ev}} }
    }
    child { node[subcatpink]{Voxel Grids}
      % child { node[paperbox]{\cite{liu2022edflow}, \cite{wang2024event}} }
    }
    child { node[subcatpink]{Time Retraining Frames}
      % child { node[paperbox]{\cite{zhou2018semi}, \cite{mitrokhin2018event}} }
    }
    child { node[subcatpink]{Spiking Representations}
      % child { node[paperbox]{\cite{zhang2024spiking}, \cite{stewart2022virtual}} }
    }
  }
  child { node[categoryred]{Datasets}
    child { node[subcatred]{Event-only}
      % child { node[paperbox]{\cite{chen2025event}, \cite{magrini2025ev}, \cite{wang2024event}} }
    }
    child { node[subcatred]{Multimodal}
      % child { node[paperbox]{ \cite{magrini2025fred}, \cite{magrini2024neuromorphic}, \cite{mandula2024towards}} }
    }
    child { node[subcatred]{Simulated}
      % child { node[paperbox]{ \cite{bhattacharya2025monocular},  \cite{carrio2018drone},  \cite{rebecq2018esim}} }
    }
  };

\end{tikzpicture}
}
\caption{\textbf{Drone Sensing Taxonomy.} We group the related works taken into account and extrapolate a precise taxonomy.}
\label{fig:taxonomy}
\end{figure}

% The papers of this survey are organized and classified as shown in Fig.\ref{fig:taxonomy}.
% \section{Related Works}
% \todo{Ha senso una sezione related in una survey? Sono 8 pagine di related}
% Numerous survey studies have recently focused on UAV related tasks. In \cite{park2021survey} a comprehensive overview of counter-drone systems is presented and broken down into the necessary tech for detection, identification and neutralization. \cite{lykou2020survey} focuses instead on drone threats in civilian airspaces and the sensors used to detect and identify \textit{rogue} drones. More recently \cite{seidaliyeva2023survey,kashi2023survey} provided a holistic state of the art review of drone detection technologies across several different modalities (radar, RF, acoustic and vision based), while \cite{liu2024survey,wang2024survey} review key challenges exclusively over vision based systems. Finally \cite{xu2025survey,semenyuk2025advances,dong2025securing} present recent advances in the use of multiple modalities, sensor fusion techniques, and benchmark datasets developed to address the detection and identification of drones. While a lot of literature covers the usage and methodology of drone detection, there persist a gap regarding the specific domain of event camera based approaches. We cover this by presenting the first (to the best of our knowledge) neuromorphic centered survey for drone detection.

\section{Drone Recognition Across Domains}
Recognition of flying objects generally falls into two main categories: natural entities like birds and insects, and artificial ones such as drones, flying vehicles and, generally speaking, Unmanned Aerial Vehicles (UAVs).
Among these, drone recognition has attracted significant attention due to growing security concerns, including risks related to unauthorized surveillance, airspace violations, and breaches of restricted zones. 
%Nonetheless, these categories often overlap and are closely interrelated. 

Significant effort has been dedicated to improving the detection of unidentified flying objects across multiple sensor domains. The most common approach relies on the RGB domain, leveraging the maturity, low cost, and high quality of standard cameras \cite{ganti2016implementation, rozantsev2015flying, hu2017detection, birch2017counter, wu2017vision, zsedrovits2011collision, mistry2023Drone,coluccia2021drone, dadboud2025drift}. Research in this area includes methods for detecting flying objects from a single moving camera \cite{birch2017counter} or tracking small unmanned aerial systems (UAS) \cite{rozantsev2015flying}. Others use visible camera systems \cite{hu2017detection}, develop real-time aerial localization and tracking systems \cite{wu2017vision}, and enable collision avoidance through visual detection \cite{zsedrovits2011collision}.
Beyond standard visual light, other optical sensors are also employed. Thermal cameras are used for night-time UAV detection~\cite{andravsi2017night}, while Short-Wave Infrared (SWIR) cameras provide robust detection capabilities in both day and night scenarios \cite{birch2017counter}. Some studies adopt alternative optical bands, such as SWIR~\cite{muller2017robust, christnacher2016optical} or combine multiple bands, such as VIS, SWIR, MWIR, and LWIR for a more comprehensive evaluation~\cite{ganti2016implementation}. Further diversifying optical methods, some works employ depth maps~\cite{carrio2018drone} or LiDAR sensors for detect-and-avoid applications~\cite{de2016flight}. Non-optical methods are also common. Radar systems, such as a 35 GHz FMCW system, are used for drone detection~\cite{drozdowicz201635}, and cost-effective Radio Frequency (RF) based detection methods have been investigated~\cite{nguyen2016investigating}. Finally, acoustic sensors can detect drones by identifying their sound signatures~\cite{mezei2015drone, christnacher2016optical}, a technique that is sometimes paired with optical detection methods~\cite{christnacher2016optical}.

%Neuromorphic approaches have been recently proposed to effectively address several facets of drone recognition. The problem has been tackled from different angles, including traditional computer vision approaches such as detection and tracking, but also introducing novel approaches that leverage the capabilities of event cameras.
%In the following, we formalize the task of drone detection with an emphasis on event cameras. Then, we will detail the challenges related to the task and the main causes and discuss event-based solutions that have been proposed in the literature.

%We next formalize the drone-detection problem in Sec. \ref{sec:taskschallenges} and analyze how event-based sensing addresses these challenges.

\section{Event-based Drone Sensing}
\label{sec:taskschallenges}

Event-based drone sensing can be declined according to several subtasks and can be addressed in different ways depending on how the event stream is represented. We summarize the key distinctions in Fig. \ref{fig:taxonomy}. In the following, we provide a formal definition of event streams and drone sensing from two different viewpoints: drone detection and drone fencing.
Let $\mathcal{E}$ denote a stream of events, defined as
\[
\mathcal{E} = \{e_i = (x_i, y_i, p_i, t_i)\}_{i \in \mathbb{N}},
\]
where $x_i \in [0, W-1]$ and $y_i \in [0, H-1]$ represent the spatial coordinates of the event $e_i$, $p_i \in \{0, 1\}$ denotes its polarity, and $t_i \in [0, \infty)$ is its timestamp.

\paragraph{Drone Detection}
The objective of the drone detection task is to train a detector $\mathcal{D}(\mathcal{I}_t)$ that produces a set of predicted bounding boxes
\[
\hat{B} = \{\hat{b}_k = (\hat{x}_k, \hat{y}_k, \hat{w}_k, \hat{h}_k, \hat{t}_k)\}_{k \in \mathbb{N}},
\]
where each bounding box is defined by its top-left corner $(\hat{x}_k, \hat{y}_k)$, its width $\hat{w}_k$, height $\hat{h}_k$, and associated timestamp $\hat{t}_k$.
The detector operates on the input $\mathcal{I}_t$, which can be either the event stream, but also a combination of the event stream $\mathcal{E}$ and a corresponding RGB stream $\mathcal{V}$, collected over a temporal window of duration $\Delta$ ending at detection time $t$. All detections at time $t$ must be based solely on data available up to time $t$, without access to any future information. 

\paragraph{Drone Fencing}
Differently from drone detection, the goal of drone fencing is not the precise localization of a drone within a scene, but rather to determine its passage or mere presence within a predefined surveilled area.
% In the drone fencing task, the objective is to train an identifier $\mathcal{F}(\mathcal{I})$ that produces a binary output $\mathcal{P}_{t}$ at any given time $t$, where we indicate with $Dr_p \in \{0,1\}$ the drone presence in the scene:
% \[
% \mathcal{P}_t = \begin{cases}
%     1 & \text{if } Dr_p \text{ is true at time } t \\
%     0 & \text{otherwise}
% \end{cases}
% \]
% for $t_i \in [0, \infty)$.
In the drone fencing task, the objective is to train an identifier $\mathcal{F}(\mathcal{I})$ that produces a binary output $\mathcal{P}_{t}$ at any given time $t$, indicating the drone presence in the scene.

\paragraph{Challenges of Drone Detection} 
The steep increment in usage and diffusion of drones in the last years has dramatically increased the related challenges for their detection. This can be linked to 4 main causes, that affect event-based drone detection: 
\begin{itemize}
    \item \textit{Variety}: modern drones are characterized by a notable variety of shape, dimensions and goals, making detection based solely on appearance hard to generalize in the wild;
    \item \textit{High speed}: the speed range in drones is notable and is not trivial to retrieve information at high fps;
    \item \textit{Extreme scenarios}: the conditions in which drone detection is necessary covers also extreme weather scenarios, low light environments, and harsh meteorological conditions (eg. rain, snow) which are challenging for event cameras too~\cite{magrini2025fred, chen2025event};
    \item \textit{Distractors and concurrent objects}: The operational environments often contain numerous other flying objects such as birds, other drones, airplanes, and even insects~\cite{coluccia2021drone,magrini2025ev}, which can act as distractors and confuse detection systems. 
\end{itemize}

\section{Event-based Drone Detection}

\begin{figure*}[t]
    \centering
    \begin{subfigure}[t]{0.21\textwidth}
        \includegraphics[width=\linewidth]{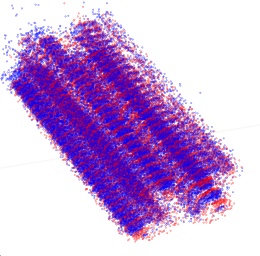}
        \caption{Event point cloud}
        \label{fig:sub1}
    \end{subfigure}
    \hfill
    \begin{subfigure}[t]{0.21\textwidth}
        \includegraphics[width=\linewidth]{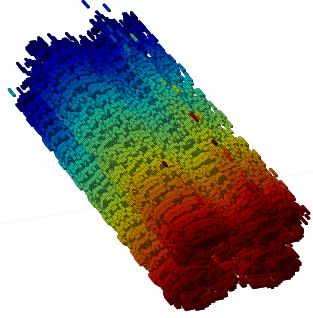}
        \caption{Event voxel grid}
        \label{fig:sub2}
    \end{subfigure}
    \hfill
    % \\
    % ~
    % \\
    \begin{subfigure}[t]{0.21\textwidth}
        \includegraphics[width=\linewidth,trim={0 5pt 0 5pt},clip]{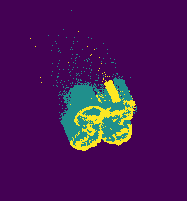}
        \caption{Time accumulation frame}
        \label{fig:sub3}
    \end{subfigure}
    \hfill
    \begin{subfigure}[t]{0.21\textwidth}
        \includegraphics[width=\linewidth,trim={0 5pt 0 5pt},clip]{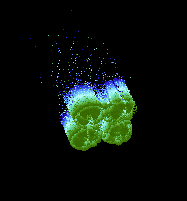}
        \caption{Time Surface frame}
        \label{fig:sub4}
    \end{subfigure}
    \caption{\textbf{Event data representations.} Comparison of different views of the event  data for a drone. Samples from the FRED dataset \cite{magrini2025fred}.}
    \label{fig:three_views}
\end{figure*}

%\todo{Qui descriverei i vari metodi proposti in letteratura. Ho ripreso quella che era la sezione metodologies ma la cambierei un po' mettendo il focus sui metodi, non su cosa cosa sono i vari approcci point/frame/ecc. Terniamo pure quello che c'è se ci torna comoda questa struttura ma riempiamo di citazioni e brevi spiegazioni/confronti di metodi per droni, tralasciando roba generica event. In generale prenderei la mega lista di citazioni in Sezione 1 (che poi si ridurrà/eliminerà) e smisterei i lavori nelle varie sottosezioni trovando un filo logico per raccontarli.}
% We classify existing event-based detectors by the representation fed to the network (point-cloud, frame, voxel, or time surface) as illustrated in Fig. \ref{fig:three_views}.

% \todo{Fed: scusate ho lasciato questo pezzo in uno stato un po' inconsistente. Proporrei di fiammarlo e di fare riferimento a Fig 1 nella intro.}

% \begin{itemize}
%     \item \textbf{4D point cloud}: thanks to the event data the continual nature, where height, width, time and polarity are the point cloud axis;
%     \item \textbf{Event Frames}: accumulating the events along the temporal axis for a given $\Delta_{t}$ results in a frame comparable to rgb or alternative domain frames;
%     \item \textbf{Voxels}: integrating event point clouds into a discrete representation in a voxel grid;
%     \item \textbf{Time-retaining Frames}: particular frame-shaped representation that integrates temporal encoding to retain more information instead of discarding the time granularity of event cameras. 
% \end{itemize}

A significant variety of event-based drone detectors is present in the literature. Their characteristics mainly depend on the nature of the data and the way in which the data is represented: on the one hand, the availability of other modalities in addition to the event stream requires the model to handle multiple parallel streams; on the other hand, different ways of representing events may require different types of architecture to process them. A comparison of common event representations is shown in Fig. \ref{fig:three_views}, where we depict a moving drone as an event point cloud, a voxel grid, or as two frame-based representations.
% In fact, event camera data can be seen as a multitude of representations, with the most common being point cloud based, event frame and voxels, for which we show a comparison in Fig. \ref{fig:three_views}.
% We start analyzing the drone based detection methods based on the used representation.

\paragraph{Frame Based} 
In contrast to the native, asynchronous nature of event data, frame-based approaches transform streams of events into dense, temporally integrated representations that mimic traditional image formats. This conversion enables the direct application of established computer vision algorithms and architectures, which require structured inputs and are difficult to adapt to irregular, sparse data.

Several recent works on drone detection with event cameras \cite{eldeborg2024drone, magrini2024neuromorphic, magrini2025fred, mandula2024towards, zundel2024bimodal} rely on frame-like representations obtained by accumulating events over a fixed temporal window $\Delta t$.
For instance, Mandula et al. \cite{mandula2024towards} accumulate events within constant intervals to produce two-channel representations: one for positive and one for negative brightness changes, which are then processed by a CNN-based detector. The authors also exploit this representation to spatially align event and RGB frames for annotation transfer.
Similarly, Magrini et al.~\cite{magrini2024neuromorphic} propose a multimodal detection pipeline in which event frames are fused with RGB inputs in a network based on Detr~\cite{carion2020end}. In this setup, events are grouped in intervals of $\Delta t = 1/F$, where F is the RGB frame rate. In Magrini et al.~\cite{magrini2025fred}, synchronized RGB and event frames are used for detection, tracking, and trajectory forecasting of UAVs in challenging real-world conditions.

In Eldeborg et al.~\cite{eldeborg2024drone}, a Spiking Neural Network (SNN) and an Artificial Neural Network (ANN) are trained using artificial event frames. The input to both models is a 128×128 frame with two polarity channels, obtained either by aggregating events within $50ms$ (ANN) or by feeding them sequentially in 1ms slices (SNN). In Zundel et al.~\cite{zundel2024bimodal}, event frames are fed into a YOLO-based detector~\cite{redmon2016you}, and DBSCAN \cite{ester1996density} is used for post-processing to identify and track drone clusters.

These studies demonstrate that frame-based event representations can be effectively integrated into conventional vision pipelines. They enable low-latency inference, efficient implementation on existing hardware, and compatibility with widely used neural architectures such as ResNet~\cite{he2016deep}, MobileNet~\cite{howard2017mobilenets}, and YOLO. Moreover, frame-based inputs simplify data augmentation, training routines, and multimodal fusion.
% However, this approach introduces important trade-offs. By aggregating events over a fixed time window, fine-grained temporal information is inevitably lost, potentially blurring fast motion and missing short-lived dynamics. This is a critical limitation in drone detection scenarios where targets are small, fast, and transient. Additionally, the inherent sparsity and high temporal resolution of the original event stream which offer advantages for low-latency, high-dynamic-range perception are largely discarded \cite{gallego2020event}.
However, this approach sacrifices fine-grained temporal details by aggregating events over time, which can blur fast motion and miss brief dynamics critical for detecting small, fast drones. It also discards the original sparsity and temporal resolution from the event stream.

\paragraph{Point Cloud Based}
Event camera streams can be seen as time-continuous point clouds, thanks to the extreme temporal granularity offered by the neuromorphic sensor. At the same time, all the information is retained. Yet the different nature of the point cloud topology and semantics, when compared to typical 3D point clouds, makes the usage of standard point-based methods non-trivial.
% A particular challenge is given by the different nature in the point cloud topology and semantic when compared to other tasks point clouds: this is given by the strict notion of time forming the 4th dimension in the event point cloud, usually separated in multiple point clouds for the other domains (eg. a temporal point cloud for action recognition will be formed by a temporally ordered sequences of 3D point clouds, while in the event camera time is not trivially separable from the single event). 
In the context of event-based tiny object detection, and specifically small or distant drones, Chen et al. \cite{chen2025event} propose the Event-based Sparse Segmentation Network (EV-SpSegNet), a U-shaped architecture designed for direct processing of sparse event point clouds. A key component is the Grouped Dilated Sparse Convolution Attention (GDSCA) module, which leverages sparse convolutions and patch attention to efficiently capture multi-scale local and long-range spatiotemporal contextual features, critical for identifying the continuous curve-like trajectories of small moving targets. 
The key idea is that the movement described by an object in the 3D event point cloud is distinctive of its nature and can therefore be used for object detection by first segmenting the point cloud. 
Complementing this, the novel Spatiotemporal Correlation (STC) Loss is introduced to specifically promote event continuity and suppress isolated noise, further enhancing the network's ability to segment tiny objects in event streams.
Another approach, presented in Magrini et al.~\cite{magrini2025ev}, is based on the classification of algorithmically detected objects in the event stream. 
In this case, the task is to correctly classify multiple classes of unidentified flying objects, specifically drones, birds and insects, using a point cloud-based network. Specifically, the usage of PointNet \cite{qi2017pointnet} and PointNet++ \cite{qi2017pointnet++} demonstrates a strong capability in distinguishing these classes by the implicit nature of their movements in an event point cloud.

\paragraph{Voxels}
Voxel grid representations \cite{zhu2019unsupervised} discretize the spatio-temporal volume of events into a regular 3D structure. The 2D sensor plane is extended along the temporal axis, which is partitioned into time bins. Each voxel encodes the number of events occurring at a given pixel location during a specific temporal interval. 
This representation captures temporal dynamics more explicitly than accumulated frames, while preserving a dense and regular structure compatible with standard deep learning architectures.
Liu et al. \cite{liu2022edflow} exploit voxel representations to compute event-driven optical flow. While not specific to drone detection, the voxel-based representation enables real-time motion analysis \cite{zhu2018multivehicle}, a key component for detecting fast-moving objects. 
%They also demonstrate superior performance accurate on MVSEC  drone and driving optical flow benchmarking sequence.
Wang et al.~\cite{wang2024event} present a more direct application, where EventVOT introduces a high-resolution benchmark and voxel-based baseline for object tracking, including UAVs. In this work, the voxel grid captures fine temporal motion information, and its regular shape facilitates the fusion with RGB modalities.

% Voxel-based methods strike a balance between temporal resolution and structural regularity. Compared to frame-based approaches, they retain more temporal detail and are less susceptible to motion blur. Compared to raw event clouds, they are more computationally tractable and can be processed using dense convolutional networks.

% However, this comes at the cost of increased memory usage and sensitivity to discretization parameters. The choice of bin size and number of temporal slices involves a trade-off: too few bins lead to loss of temporal information, while too many increase sparsity and computational burden. Additionally, voxelization introduces quantization noise, potentially discarding subtle timing information preserved in raw event streams..

\paragraph{Time-retaining Frames}
While conventional frame representations aggregate events into static histograms over a fixed temporal window $\Delta t$, time-aware frame representations aim to retain the temporal structure of the event stream within a frame-like format. These encodings not only capture spatial event density but also preserve the relative timing of individual events, enabling networks to exploit dynamic motion cues.
A classic example of time-retaining approach is Time Surface \cite{sironi2018hats}, where each pixel value encodes the timestamp of the most recent event.
Similarly, Temporal Binary Representation (TBR) \cite{innocenti2021temporal} and its spiking variant SpikeTBR \cite{magrini2025spike} encode temporal information into the pixel values of frame representations.
%Beyond time surfaces, other temporally sensitive encodings have been proposed in different domains. Temporal Binary Representation (TBR) \cite{innocenti2021temporal} and its spiking variant SpikeTBR \cite{magrini2025spike} have shown promise in gesture recognition and may be adaptable to UAV detection.
%\cite{zhou2018semi} employ time surfaces generated from stereo event data to reconstruct 3D geometry in dynamic scenes. Though not focused on drone detection, the work showcases how time-aware representations can enhance the understanding of motion and structure key components in fast moving object detection. 
Mitrokhin et al.\cite{mitrokhin2018event} leverage a motion-compensated time image to segment moving objects from background motion. The representation encodes per-pixel timing information, allowing the system to disentangle drone trajectories from camera-induced motion. 
% Time-aware representations offer a compelling trade-off between temporal fidelity and architectural compatibility. They maintain continuity in motion representation and can be processed efficiently by standard CNNs. This is especially beneficial when targets such as drones exhibit rapid, dynamic behavior.

% Despite their strengths, time-aware representations require careful tuning of temporal decay or update functions. They often discard polarity or event density information, which can be useful in high-speed or cluttered scenes. Nonetheless, their ability to preserve short-term motion history while remaining computationally efficient makes them a valuable alternative to both frame and voxel-based approaches in drone detection tasks.

\paragraph{End-to-end Neuromorphic approaches}
This scenario involves a complete pipeline of neuromorphic sensing, processing, and computing. Unlike hybrid approaches, where event data is converted into intermediate representations to interface with conventional deep learning models, end-to-end neuromorphic systems maintain the spike-based nature of the signal throughout the entire processing chain. This includes both Spiking Neural Network (SNN) architectures and dedicated neuromorphic hardware, such as Intel's Loihi \cite{davies2018loihi} or SynSense's Speck \cite{Yao_2024}, which are designed to exploit the sparse, event-driven characteristics of the input.
These systems offer unmatched energy efficiency and latency, making them especially suitable for edge deployments in scenarios like drone detection, where power and bandwidth constraints are critical. 
% The training of SNNs remains a significant challenge, due to the non-differentiable nature of spikes; however, recent advances in surrogate gradient learning \cite{neftci2019surrogate} and conversion-based methods (e.g., ANN-to-SNN conversion \cite{rueckauer2017conversion}) have enabled increasingly competitive performance in real-world tasks.
% Moreover, the temporal coding of spikes provides a natural fit for modeling fine-grained time-dependent phenomena, with temporal precision beyond that of frame-based systems. Methods such as time-to-first-spike coding \cite{mostafa2017supervised}, latency coding, and phase coding have been explored to encode information efficiently and biologically plausibly. When paired with event-based sensors, end-to-end neuromorphic systems enable truly asynchronous and continuous processing, closely mimicking biological perception systems. This paradigm represents a significant shift from traditional frame-based pipelines, favoring event sparsity, low latency, and biological plausibility over data redundancy and batch processing.
In Kirkland et al.~\cite{kirkland2019uav}, the authors propose a low-power UAV detection system based on a retina-inspired neuromorphic vision sensor, leveraging a deep convolutional Spiking Neural Network trained using Spike-Timing Dependent Plasticity (STDP). The system is designed to detect drones by processing asynchronous event streams in real time, demonstrating the potential of STDP-trained SNNs for efficient aerial threat detection in edge scenarios.
Another significant advantage of end-to-end neuromorphic pipelines is the extremely low power consumption; in Eldeborg et al.~\cite{eldeborg2024drone} the authors present a low-power drone detection system utilizing the Speck processor and embedded event-based camera to establish a virtual tripwire, with a Spiking neural network tasked with classifying the presence of the drone in the scene (drone fencing task, see Sec. \ref{sec:taskschallenges}). 
In Zhang et al.\cite{zhang2024spiking}, the authors propose a low-power UAV object detection system based on event cameras, leveraging a Spiking Swin Transformer architecture that combines spiking neurons with hierarchical transformer features. In this case, no deployment on a real neuromorphic chip has been used.

\section{Beyond Neuromorphic Drone Detection}

\paragraph{Drone Tracking}

A natural extension of the drone detection task is drone tracking. In the drone tracking task, drone instances must be spatio-temporally identified across the entirety of a stream of events.
Several event-based general-purpose tracking approaches have been proposed in the literature. Early works adapted traditional computer vision techniques, such as combining Discriminative Correlation Filters (DCF) with Convolutional Neural Network (CNN) features by first converting event streams into frame-like representations \cite{li2019robust}. SiamEvent \cite{chae2021siamevent} shifted the focus to learning an edge-aware similarity metric more aligned with the information content of events. More recently, spiking neural networks have also been used for object tracking~\cite{zhang2022spiking}. 

The first work to address drone tracking~\cite{mitrokhin2018event} proposed the EED dataset, which contained small drones among other objects. The task was addressed by developing an unconstrained motion segmentation model that estimated the effects of 3D camera motion from the event stream.
A larger benchmark for object tracking including also UAVs, EventVOT \cite{wang2024event}, was recently proposed. The authors also present HDETrack, a tracking framework that leverages knowledge distillation, training a lightweight, event-only student tracker to mimic a powerful, multimodal RGB-Event teacher network, achieving high performance with low latency.
Recently, some works have focused solely on tracking drones.
Iaboni et al.~\cite{iaboni2022event} proposed a low-cost motion capture system that uses an event camera to track multiple quadrotors in real-time. The method combines YOLOv5 with KD-tree tracking, trained on synthetic event frames.
End-to-end models for real-time neuromorphic detection and tracking of UAVs have also been proposed~\cite{skogsberg2024event}: an asynchronous, event-by-event method, combines corner tracking with clustering to identify targets, a design choice aimed at minimizing computational load. A more traditional model is also proposed in the same work, where the event stream is first accumulated into event frames.

\begin{figure}[t]
    \centering
    \includegraphics[width=\linewidth]{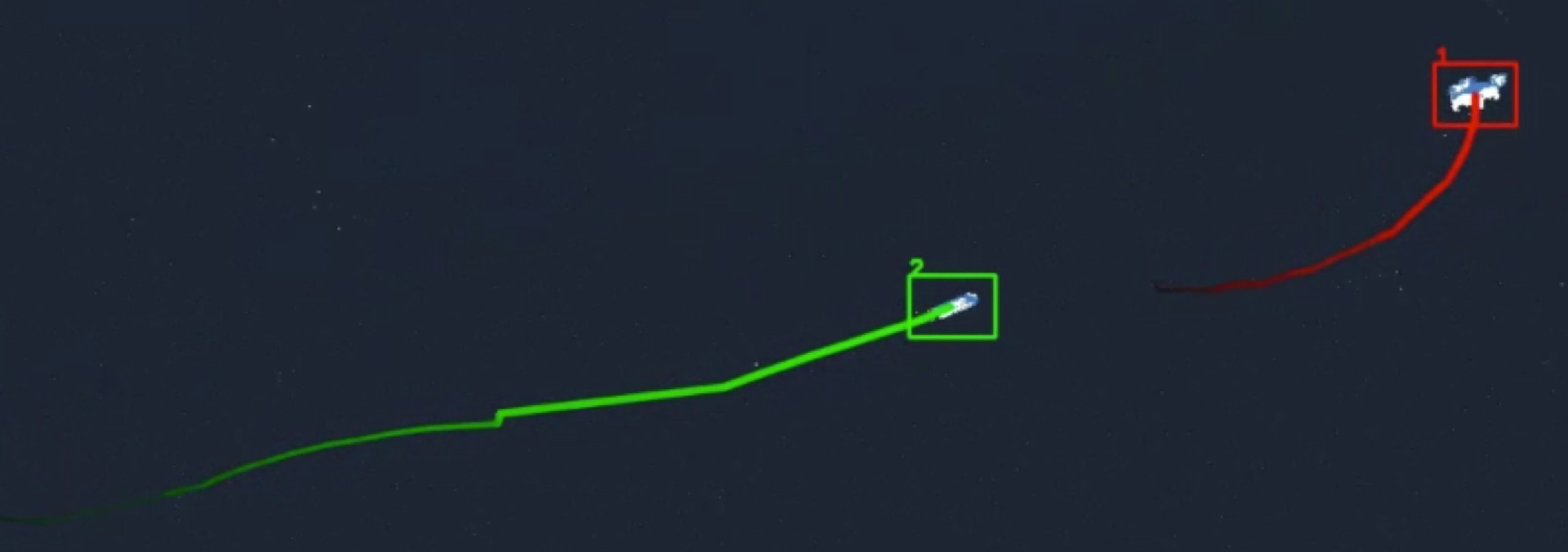}
    \caption{\textbf{Drone tracking task.} Drone tracking annotations taken from the FRED dataset \cite{magrini2025fred}. Different colors correspond to different drones.}
    \label{fig:tracking}
\end{figure}

The authors of the FRED dataset \cite{magrini2025fred} instead introduced a larger-scale benchmark for drone detection and tracking, better formalizing the task and proposing several tracking-by-detection baselines, combining object detections such as YOLO~\cite{khanam2024yolov11}, RT-DETR~\cite{zhao2024detrs} and Faster-RCNN~\cite{girshick2015fast} with the state-of-the-art tracker ByteTrack~\cite{zhang2022bytetrack}. An RGB-Event multimodal tracker is also proposed, based on ER-DETR~\cite{magrini2024neuromorphic}. MOTA (Multi-Object Tracking Accuracy), IDF1 (Identity F1 score), ID Switch, Precision and Recall are used to evaluate the trackers. An example is shown in Fig.~\ref{fig:tracking}.

\paragraph{Drone Forecasting}

\begin{figure}[t]
    \centering
    \includegraphics[width=\linewidth]{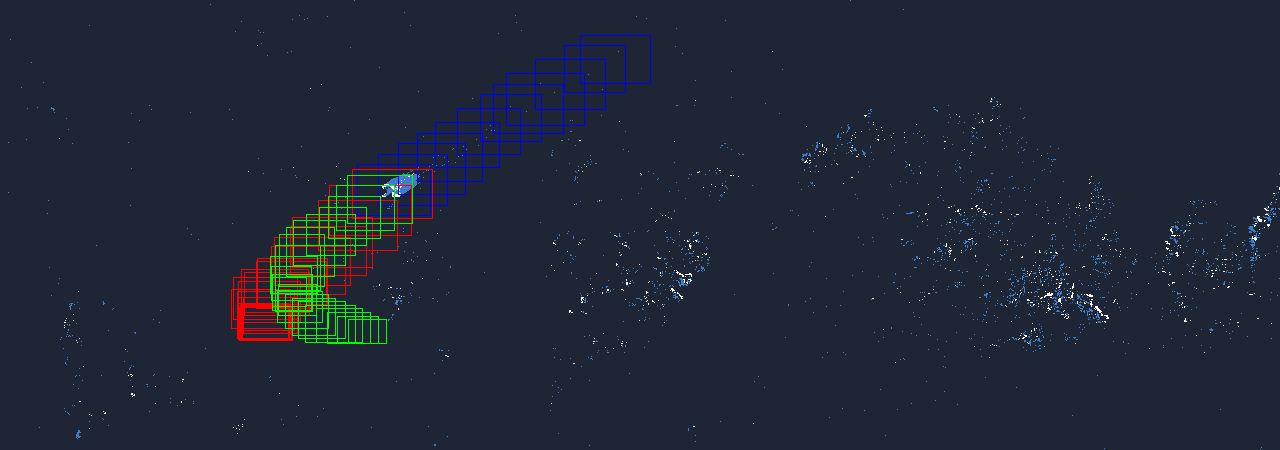}
    \caption{\textbf{Drone forecasting task}. Annotation for the forecasting task from FRED \cite{magrini2025fred}. Blue: past; Green: GT; Red: prediction.}
    \label{fig:forecasting}
\end{figure}

Forecasting the position of fast-moving objects is an extremely challenging task. However, the benefits of event cameras have been demonstrated in the literature~\cite{monforte2023fast}. Despite this, only a few works exist on forecasting the trajectories of UAVs with event cameras, which will likely become a fundamental application in the near future.
Liang et al. Liang et al.~\cite{liang2025label} proposes an unsupervised approach that utilizes raw LiDAR point clouds to extract drone trajectories and aligns them with event camera images through motion consistency to generate pseudo-labels. By combining kinematic estimation with a visual Mamba neural network in a self-supervised manner, the method predicts future drone trajectories, outperforming supervised image-only and audio-visual baselines in long-horizon predictions.
Magrini et al.~\cite{magrini2025fred} instead propose the only publicly available benchmark for event-based drone forecasting. The authors present blind references based on LSTMs or transformers that observe only drone coordinates and show that event-based models can largely improve the performance. Interestingly, event-based models consistently outperform RGB-based ones and the proposed multimodal approach outperforms unimodal models for short-term predictions. An example is shown in Fig. \ref{fig:forecasting}.

\paragraph{Propeller Blade Analysis}
% Propeller speed estimation \cite{zhao2022high, zhao2023ev}
% Angular speed measurement \cite{azevedo2025multiple}
% Detecting Drones by finding propellers \cite{sanket2021evpropnet}
% Propeller detection \cite{spetlik2025efficient}

\begin{figure}[t]
    \centering
    \includegraphics[width=\linewidth]{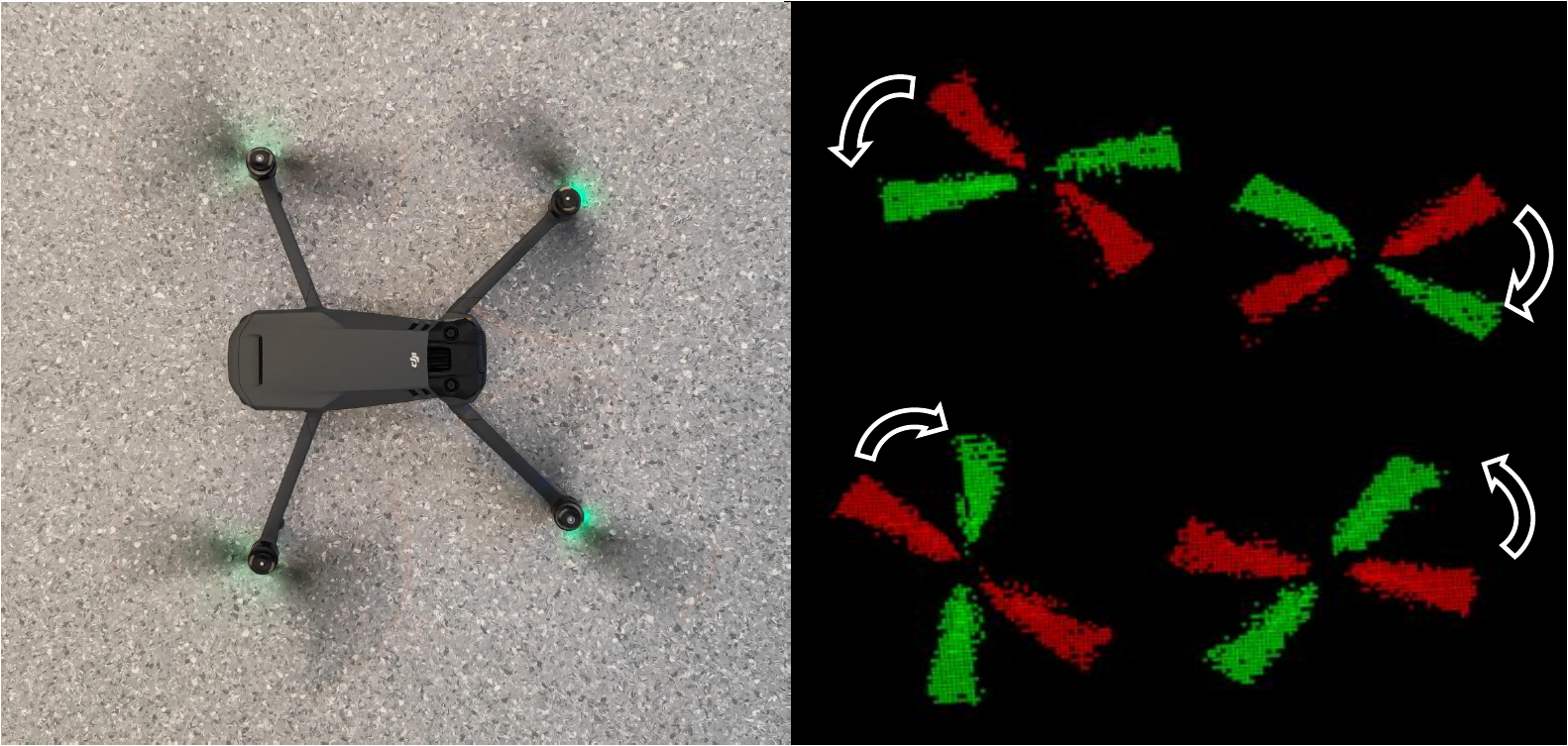}
    \caption{\textbf{Propeller Blade Analysis.} Drone propeller blades observed with an event camera. Image taken from \cite{zhao2022high}.}
    \label{fig:blade}
\end{figure}

A highly effective method for drone identification using event cameras is the analysis of the visual signature generated by propellers. As can be seen in Fig.~\ref{fig:blade}, the rapid rotation of blades, which causes motion blur for standard cameras, creates a rich, high-frequency spatio-temporal pattern perfectly suited for the microsecond temporal resolution of event sensors \cite{azevedo2025multiple, sanket2021evpropnet, spetlik2025efficient}. These sensors leverage their high temporal resolution, high dynamic range (HDR), and sparse, event-driven output to capture the periodic brightness changes from the blades, effectively filtering out the static background \cite{sanket2021evpropnet, spetlik2025efficient,hoseini2017propeller}. This unique signature serves as a robust feature to distinguish drones from other objects and effectively acts as an uncooperative, natural fiducial marker for detection and tracking \cite{sanket2021evpropnet}.

Several methodologies have been developed to analyze this propeller signature. Frequency-domain techniques treat the event stream as a periodic signal to extract its core frequency components. This analysis is conceptually analogous to the use of micro-Doppler signatures in radar systems for drone identification \cite{drozdowicz201635, eldeborg2024drone}. More advanced probabilistic methods offer greater robustness. One of these approaches models local event arrivals as Poisson processes to statistically distinguish the periodic bursts from propeller blades against background noise in real-time \cite{spetlik2025efficient}. Finally, learning-based approaches have leveraged deep neural networks. The seminal work, EVPropNet, trained a CNN exclusively on a large-scale synthetic dataset of propeller events \cite{sanket2021evpropnet}. This network transfers directly to real-world scenarios without retraining, enabling complex on-drone applications like autonomous following and mid-air landing on unmarked drones, shifting the paradigm from passive surveillance to active robotic interaction.

\begin{table*}[th]
    \centering
    \renewcommand{\arraystretch}{1.2}
    \resizebox{\textwidth}{!}{
    \begin{tabular}{l|c|c|c|c|c|c}
    \hline
        \textbf{Dataset} & \textbf{Resolution} & \textbf{RGB / Event} & \textbf{Duration} & \textbf{Drone-Centric} &  \textbf{Drone Types} & \textbf{Tasks} \\
        \hline
        \textbf{Anti-UAV~\cite{jiang2021anti}}     & 1920 $\times$ 1080      & \checkmark{} / $\times$ & 4h:25m     & \checkmark{} & $\geq 6$ & Det,Track \\
        \textbf{Pawelczyk et al.~\cite{pawelczyk2020real}}  & 640$\times$480  & \checkmark{} / $\times$ &   51446 frames   & \checkmark{} & - & Det \\
        \textbf{EED~\cite{mitrokhin2018event}}  & 240 $\times$ 180 & $\times$ / \checkmark{} & $\leq$ 0h:30m  & $\times$ & - & Det, Track \\
        \textbf{VisEvent~\cite{wang2023visevent}}     & 346 $\times$ 260      & \checkmark{} / \checkmark{} & $\leq$ 5h     & $\times$ & - & Det \\
        \textbf{EventVOT~\cite{wang2024event}}        & 1280 $\times$ 720     & $\times$ / \checkmark{}     & $\leq$ 5h     & $\times$ & - & Det\\
        \textbf{F-UAV-D~\cite{mandula2024towards}}    & 1280 $\times$ 720     & \checkmark{} / \checkmark{} & 0h:30m        & \checkmark{} & 2 & Det\\
        \textbf{Ev-UAV~\cite{chen2025event}}          & 346 $\times$ 240      & $\times$ / \checkmark{}     & 0h:15m        & \checkmark{} & - & Det,Track\\
        \textbf{Ev-Flying~\cite{magrini2025ev}}   & 1280 $\times$ 720     & $\times$ / \checkmark{}     & 1h:07m        & $\times$ & 1 &  Det,Track\\
        \textbf{NeRDD~\cite{magrini2024neuromorphic}} & 1280 $\times$ 720     & \checkmark{} / \checkmark{} & 3h:30m        & \checkmark{} & 2 & Det\\
        \textbf{CRSOT~\cite{zhu2024crsot}} & 1280 $\times$ 720     & \checkmark{} / \checkmark{} & 2h:50m        & \checkmark{} & - & Det,Track\\
        \textbf{FRED~\cite{magrini2025fred}}          & 1280 $\times$ 720     & \checkmark{} / \checkmark{} & 7h:07m        & \checkmark{} & 5 & Det,Track,Forecast\\
        \hline
    \end{tabular}
    }
    \caption{\textbf{Comparison of event-based drone datasets.} RGB/Event: availability of modalities. Drone-centric: whether the drone is the primary subject. Task: which tasks the dataset is aimed at (Detection, Tracking, Forecasting)}
    \label{tab:datasets}
\end{table*}

\setlength{\tabcolsep}{4pt} % Reduce column separation
\renewcommand{\arraystretch}{1.5} % Increase row height slightly for vertical centering
\newcommand*{\thead}[1]{\multicolumn{1}{|c|}{\bfseries #1}} % Define \thead for centered headers

These analysis methods enable precise quantitative measurements of a drone's operational state. The field of event-based tachometry focuses on estimating propeller rotational speed with high fidelity. Systems like EV-Tach have demonstrated performance comparable to commercial laser tachometers, with relative errors as low as 0.03\%, while functioning in difficult lighting conditions \cite{zhang2024spiking, zhao2023ev}. Other methods, like EB-ASM, also achieve high accuracy and have demonstrated the crucial capability of measuring the speeds of multiple propellers simultaneously with a single camera, which is essential for monitoring multi-rotor drones \cite{azevedo2025multiple}. Beyond speed, the analysis can be extended to full state estimation. For instance, by fitting an ellipse to the aggregated event cloud generated by a propeller, it is possible to derive the drone's pitch and roll angles, providing valuable attitude information \cite{spetlik2025efficient}. The evolution of these applications shows a clear progression from simple detection to sophisticated tachometry and state estimation, establishing propeller signature analysis as a key technology for both remote surveillance and advanced autonomous robotics.

In the work by Stewart et al~\cite{stewart2021drone, stewart2022virtual}, the authors present a virtual fence system for drone detection based on event cameras. The approach leverages the unique rotational signature of drone propellers, captured as periodic patterns in the event stream. By computing temporal histograms and applying spectral analysis, the system detects drones by identifying characteristic frequency peaks associated with propeller motion. 
A substantial improvement was then presented \cite{stewart2023using}, in which the propeller analysis is used directly for the drone spatial detection instead of only presence detection.
These methods are purely algorithmic and need no neural network or training data.

\section{Datasets}
Before delving deep into the existing datasets and their characteristics, we need to make a further distinction. 
Since event cameras alone provide for many advantages not present in other sensors, many works in the literature focus on data gathered by this sensor only. 
On the other side, the usage of multiple sensors concurrently presents the advantage of better reliability, especially in the presence of challenging scenarios and a more complete coverage of the scene by using complementary domains (e.g., the RGB domain can retain color information that may be crucial for distinction between drones and other kinds of flying objects).
Tab. \ref{tab:datasets} provides an overview of event-based drone sensing datasets.

\paragraph{Event Only Datasets}
Many methods rely solely on event data due to its rich temporal and spatial content.
Single-modality data is easier to collect, requiring only labeling and no sensor synchronization.
We can make a further distinction: drone-centered datasets and drone-containing datasets. 
This difference pertains to the simple inclusion of drone instances in a broader dataset (drone-containing) versus datasets that focus on drone detection as their primary objective (drone-centered).
A collection of 10,000 event frames, among the first drone-centered event based dataset, was first presented by Iaboni et al.~\cite{iaboni2022event}. Here, the dataset focuses on multi-quadrotor localization and tracking. The system uses a ceiling-mounted event camera to monitor quadrotors flying indoors and outdoors. The dataset captures multiple quadrotors (up to 6) in motion across various lighting conditions, flight speeds, and altitudes.
In \cite{chen2025event}, the authors present Ev-UAV, a dataset comprising 147 sequences and over 2.3 million event-level annotations, featuring extremely small targets. The dataset captures diverse scenarios, including urban clutter and extreme lighting conditions.
An alternative approach is presented in Da et al. \cite{da2025new}: here the authors present SFERA, a novel stereo fisheye event camera system developed for fast, omnidirectional drone detection and tracking. Two synchronized fisheye Prophesee cameras provide 360$^\circ$ spherical coverage. The dataset includes over 35 minutes of annotated recordings of drones in outdoor environments under various flight trajectories. Annotations provide per-frame drone positions projected in spherical coordinates.
Other works focus on generic object detection, with a broader focus on large quantities of classes, but also including recordings of drones \cite{wang2025object, wang2024event, mitrokhin2018event}.

\paragraph{Multi-modal Datasets}
\label{sec:multimodaldataset}
Beyond the event camera-only datasets, another subset of datasets comprises complementary domains to be used with the neuromorphic data. These kinds of approaches have a number of advantages as well as some practical disadvantages. On one hand, they provide for more resilient data to domain-specific limitations: one common event camera problem is in fact related to scenarios in which objects may remain stationary for a certain period of time, leading to a total loss of information about said objects. In this scenario, a complementary RGB camera can improve the chance of retaining information \cite{wang2020joint, tomy2022fusing}. RGB data also helps to distinguish drones from birds, insects, and other flying agents. 
Some of the major downsides of these approaches are related to the difficulty in setting up the recording station when multiple sources are present.
% Each additional sensor requires particular care when dealing with both the temporal and spatial dimensions, and while some cameras are inherently multi modal with comprised event cameras as the DAVIS346, particular sensors must be integrated with other solutions. 
A general-purpose dataset, comprising also annotation of drones, is presented in Wang et al.~\cite{wang2023visevent}, in which the authors gather 820 video pairs (RGB + event) collected across real-world scenarios featuring fast motion, low illumination, and cluttered backgrounds. The dataset focuses on tracking and adopts a low-resolution event camera (346$\times$240).
Similarly, Zhu et al.~\cite{zhu2024crsot} tackle this approach with a more diverse data scenario, with the main goal of tracking different objects at high resolution.
In Mandula et al. \cite{mandula2024towards}, the authors introduce F‑UAV‑D, a dedicated RGB-neuromorphic dataset for event‑based UAV detection. The dataset was captured using the Prophesee EVK4 (1280$\times$720) for the event stream and Sony’s IMX219 for the RGB frames. The dataset comprises approximately 30 minutes of event-frame videos featuring fast-moving drones in both controlled indoor and outdoor scenarios. 
In Magrini et al.~\cite{magrini2024neuromorphic}, the authors introduce NeRDD (Neuromorphic‑RGB Drone Detection), a multimodal benchmark dataset with spatio-temporally synchronized event (DVS) and RGB data, with over 3.5 hours of annotated drone footage collected from stationary ground cameras. 
Another similar dataset, FRED \cite{magrini2025fred}, is composed of a total of more than 7 hours of annotations for drone detection, tracking and forecasting, which includes high-resolution recordings of five drone models and challenging settings such as rain, low light, insect distractors, and indoor/outdoor scenarios.

\paragraph{Simulators and Synthetic Data}
Since data gathering is a complex and time-consuming process, especially regarding the labeling phase, another frequently common method of creating drone-related datasets is through the simulation of drone-populated environments or the generation of synthetic data. Labeling is potentially free, and scene parameters (e.g., lighting, weather, motion) are fully controllable.
% It can also be efficient resource-wise, since it does not requires the often costly equipment as the cameras and drones. 

Simulators are also frequently paired with event simulation networks, which create synthetic events from temporally upsampled RGB or grayscale images \cite{rebecq2018esim, Gehrig_2020_CVPR}.
This can be applied to both On-Drone \cite{bhattacharya2025monocular} and Off-Drone approaches. 
The main disadvantage of this approach is the limited environment realism, with risks of domain shift when compared to real recordings and scenarios.
Simulations may be slow and computationally heavy, with limited lighting and temporal realism, both critical for event data.  

Another direction for synthetic data generation is the usage of 3D environments with embedded physical properties, which can result in more realistic data and light simulation. 
Specialized tools for drone flight simulation have also been developed: Shah et al. \cite{shah2017airsim} proposed an event-based camera simulation module that takes the output from a regular simulated camera, applies a non-linear operation, subtracts successive frames, and then thresholds the outcome. For accurately capturing a rotating propeller's spectral signature, this approach would demand an exceptionally high frame rate from the simulated camera. 
In Eldeborg et al. \cite{eldeborg2024drone}, the authors propose a mix of a small portion of real neuromorphic data combined with a substantial amount of synthetic data, simulated using Blender: in this case, the goal is to simulate the rotating propellers of a UAV.

%\section{Methodologies}

\section{Conclusions}
In this paper, we addressed the critical challenge of drone detection, where traditional cameras are often hindered by motion blur and limited dynamic range. Event cameras offer a robust alternative, using their high temporal resolution, high dynamic range, and sparse data output to effectively capture small, fast-moving drones in challenging lighting conditions.
The field is advancing rapidly, with methods evolving from converting event data into frames for use in conventional neural networks to processing it in its raw point cloud form or with spiking neural networks. Research has also progressed beyond simple detection to tackle more complex tasks such as drone tracking, trajectory forecasting, and unique identification through the analysis of propeller blade signatures.
Event-based vision addresses the limits of conventional sensors, enabling efficient, low-latency counter-UAV solutions.

\section*{Acknowledgements}
This paper was partially funded by Leonardo S.p.A. and by the project "Collaborative Explainable neuro-symbolic AI for Decision Support Assistant", CAI4DSA, CUP B13C23005640006.
This work was partially supported by the Piano per lo Sviluppo della Ricerca (PSR 2023) of the University of Siena, project FEATHER: Forecasting and Estimation of Actions and Trajectories for Human–robot intERactions.

%Event cameras \cite{gallego2020event}

%Drone detection \cite{farah2025ev, cao2024eventboost, chuecaonboard, da2025new, magrini2024neuromorphic, stewart2022virtual, mitrokhin2018event, miao2025dual, shu2021small,  magrini2025fred, muller2017robust, liu2022edflow, iaboni2022event, magrini2025ev, svanstrom2022drone}

% \section{Method}
% \todo{note per eventuali esperimenti, forse si toglie tutto?}
%  YOLO event con varie codifiche:
% \begin{itemize}
%     \item Surface of Active Events (SAE) \cite{mueggler2017fast}
%     \item TBR
%     \item Polarity \cite{nguyen2019real}
%     \item Prophesee? È uguale a polarity?
%     \item VoxelGrid? Si usano 3 bin temporali?
%     \item Bai et al. \cite{bai2022accurate}
% \end{itemize}

% Ablation:
% \begin{itemize}
%     \item YOLO pretrained vs. from scratch
%     \item Variare il tempo di accumulazione
% \end{itemize}

{
    \small
    \bibliographystyle{ieeenat_fullname}
    \bibliography{main}
}

\end{document}